\pdfoutput=1
\documentclass[11pt]{article}
\usepackage[]{acl} 
\usepackage{times}
\usepackage{latexsym}
\usepackage[T1]{fontenc}
\usepackage[utf8]{inputenc}
\usepackage{microtype}
\usepackage{inconsolata}

\usepackage{geometry} 
\usepackage{color}
\usepackage{amsfonts}
\usepackage{enumitem}
\usepackage{algorithm}
\usepackage{algpseudocode}
\usepackage{amsmath}
\usepackage{graphicx}
\usepackage{hyperref}
\usepackage{tikz,lipsum,lmodern}
\usepackage[most]{tcolorbox}

\newlist{steps}{enumerate}{1}
\setlist[steps, 1]{label = \textbf{Step \arabic*:}}

\author{He Yan\textsuperscript{1}, Xinyao Hu\textsuperscript{1}, Xiangpeng Wan\textsuperscript{1}, Chengyu Huang\textsuperscript{2}, Kai Zou\textsuperscript{1}, Michael Shiqi Xu\textsuperscript{3} \\
        \textsuperscript{1}ProtagoLabs \\
        \textsuperscript{2}International Monetary Fund \\
        \textsuperscript{3}NetMind.ai \\
        {\{he.yan, xinyao.hu, xiangpeng.wan, kz\}@protagolabs.com} \\
        {chuang@imf.org} \\
        {michael@netmind.ai}
        }

\title{Inherent limitations of GPT-4 regarding spatial information}
\date{August 2023}

\begin{document}
\maketitle

\begin{abstract}
\boldmath
Despite the significant advancements in natural language processing capabilities demonstrated by large language models such as ChatGPT, their proficiency in comprehending and processing spatial information, especially within the domains of 2D and 3D route planning, remains notably underdeveloped. This paper investigates the inherent limitations of ChatGPT and similar models in spatial reasoning and navigation-related tasks, an area critical for applications ranging from autonomous vehicle guidance to assistive technologies for the visually impaired. In this paper, we introduce a novel evaluation framework complemented by a baseline dataset, meticulously crafted for this study. This dataset is structured around three key tasks: plotting spatial points, planning routes in two-dimensional (2D) spaces, and devising pathways in three-dimensional (3D) environments. We specifically developed this dataset to assess the spatial reasoning abilities of ChatGPT. Our evaluation reveals key insights into the model's capabilities and limitations in spatial understanding.
\end{abstract}

\section{Introduction}

In the real world, there are numerous problems related to spatial understanding and reasoning, such as map navigation, autonomous driving, and engineering drafting. Some spatial information can be described in natural language, which humans can understand with simple common sense. Recently, aligned Large Language Models (LLMs) like ChatGPT and Claude 2 have demonstrated remarkable abilities to follow instructions\cite{ouyang2022training}. As the capabilities of LLMs continue to improve, traditional evaluation standards have become obsolete, and new benchmarks have been proposed to assess LLMs' ability to process extensive world knowledge or reason through complex tasks. Benchmarks such as MMLU~\cite{hendrycks2020measuring}, BIG-bench~\cite{srivastava2022beyond}, and AGIBench~\cite{tang2023agibench} have brought together a variety of tasks to comprehensively evaluate the capabilities of LLMs. There are also benchmarks that assess LLMs' abilities in specific areas, such as mathematical ability~\cite{hendrycks2021measuring,cobbe2021training,welleck2021naturalproofs,drori2023dataset,yang2023leandojo}, programming skills~\cite{chen2021evaluating,zheng2023codegeex,du2023classeval,athiwaratkun2022multi}. In some benchmarks, LLMs can achieve or surpass human-level performance, but currently, there is no benchmark specifically designed to evaluate LLMs' spatial understanding and reasoning abilities.

We believe that such a benchmark is necessary as a first step towards using language models to solve complex spatial problems in the real world. In this work, we have designed three different tasks focused on assessing LLMs' spatial understanding and reasoning abilities. The three tasks are plotting spatial points, planning routes in two-dimensional (2D) spaces, and devising pathways in three-dimensional (3D) environments. For each task, we have created a set of unique, solvable problems based on specific rules. During problem generation, parameters can be adjusted to control the complexity of the problems, thereby evaluating LLMs' performance on the same task at different levels of difficulty. We have written code using traditional methods to obtain standard answers, such as employing Dijkstra's algorithm to find the shortest path for planning routes. Subsequently, we use LLMs to generate answers and employ code to automatically verify the validity and correctness of the answers produced by the LLMs.

The main contributions of this paper are summarized as follows:
\begin{itemize}
    \item Novel Evaluation Framework: The paper introduces an innovative framework for evaluating the spatial reasoning abilities of large language models like GPT-4. This framework represents a significant step in understanding and quantifying how such models deal with spatial information.
    \item Bespoke Baseline Dataset: A specially designed dataset, tailored for this study, constitutes a major contribution. This dataset focuses on three spatial tasks—plotting spatial points, 2D route planning, and 3D route development—providing a unique and targeted tool for assessing the proficiency of language models in spatial reasoning.
    \item Insights into Model Capabilities and Limitations: The paper offers critical analysis and findings on the abilities and shortcomings of GPT-4 in handling spatial information. This analysis is crucial for applications requiring spatial awareness and navigation, thereby informing future model development and adaptation in fields like autonomous vehicle navigation and assistive technologies for the visually impaired.
    
\end{itemize}

\section{Related Works}
Recent efforts to evaluate Large Language Models (LLMs) have focused on their ability to process extensive world knowledge and reasoning through complex tasks. These evaluations can be divided into reference-based and reference-free methods. Reference-free methods~\cite{wang-etal-2023-self-instruct,alpaca_eval,lfqa23,pandalm2023}, such as Chatbot Arena~\cite{zheng2023judging}, involve human evaluation of model responses, while reference-based methods typically present language models with multiple-choice questions or compare model outputs against exact answers. Notable benchmarks include the MMLU, which offers multi-domain tasks from real-world sources, and the BIG-bench, featuring 204 diverse tasks, some designed to challenge LLMs with problems they cannot fully solve. Some benchmarks are created to evaluate performance in specific domains~\cite{deng2023mind2web, wang2023decodingtrust,lin2021truthfulqa,schwettmann2023find,gandhi2023understanding}. For example, GSM8K~\cite{cobbe2021training} is aimed at elementary math problems, HumanEval~\cite{chen2021evaluating} tests programming skills, and BIRD-SQL~\cite{li2023can} focuses on converting text into SQL queries.

In previous work, tasks from the BIG-bench benchmark, such as ``navigate'' and ``geometric shape'' have been associated with spatial understanding. The ``geometric shape'' task requires models to identify shapes from SVG path elements, while ``navigate'' relies on natural language descriptions that could be resolved through mathematical calculations. The High School Math section of the MMLU benchmark includes analytic geometry problems that necessitate high school-level math and logical reasoning, which do not directly assess spatial understanding. Some studies have assessed the abilities of Large Multimodal Models (LMMs) on the Abstraction and Reasoning Corpus (ARC)~\cite{mitchell2023comparing, moskvichev2023the} and have also converted two-dimensional images into text-based grid inputs for text-only GPT-4~\cite{openai2023gpt4}. However, the text generated in these studies often lacks an intuitive understanding of the images they represent. ARC tasks primarily require models to abstract underlying rules from a few examples and generalize them to unseen situations, a challenge that goes beyond the assessment of spatial understanding and reasoning capabilities.

Our work differs from these previous approaches by focusing exclusively on LLMs' spatial understanding capabilities without the need for additional specialized knowledge. We have designed tasks with adjustable difficulty levels, ranging from simple to complex, to enable a quantitative evaluation of language models' performance in spatial reasoning. Each task is crafted to allow for a scalable increase in complexity, with quantifiable metrics for difficulty, ensuring a nuanced assessment of spatial understanding in LLMs.

\section{Experiments}

\subsection{Experiments 1 - 2D Route Planning}
\label{de:tf} 
In our 2D route planning experiments, we evaluate the spatial reasoning abilities of GPT-4 by asking GPT-4 to discover the optimal path in 2D space with obstacles through text-based prompts. Starting from the orgin point$(0, 0)$ to the arriving point($x_{max}$, $y_{max}$). Finally, we employ Dijkstra's algorithm as the shortest path benchmark. We evaluate GPT-4's ability to find the shortest path. These prompts encompass the following criteria:
\begin{itemize}
    \item Each step must take a distance of 1 unit.
    \item You must not traverse beyond obstacles.
    \item Ensure the precise determination of the starting point$(0,0)$ and arriving points($x_{max}$, $y_{max}$).
    \item The total distance cost should match that of Dijkstra's algorithm.
\end{itemize}

\subsubsection{Experiment Setup}

We need to create the grid database by adjusting the two parameters: grid dimensions and the proportion of obstacles. The database covers nine different grid sizes, starting from 3$\times$3 and going up to 11$\times$11. In each grid size, we introduce diversity by randomly adding obstacles and the obstacle ratios for each grid are 0\% (no obstacles), 5\%, 10\%, 15\%, 20\%, and 25\%. Furthermore, we generated ten sets of obstacles randomly for each obstacle ratio to control the complexity of the experiment. To maintain the experiment consistency, we exclude certain obstacle ratios for smaller grid sizes. Next, we employed the Dijkstra algorithm to obtain the correct total distance of the grid data set, and finally, we obtain all benchmarking data which contains approximately 470 sets of grid data. It is noted that we exclude the grids with no possible paths in our experiment.

Finally, we guide GPT-4 to find the optimal path in the grid database and evaluate its ability to find the shortest path.

\subsubsection{Results}

The chart illustrates GPT-4's accuracy in finding the shortest path in different map sizes and obstacle ratios. Due to the instability of GPT-4, we direct GPT-4 to find the shortest path for the map sizes with no obstacles ten times. The results reveal that as map size and obstacle ratio increase, the ability to correctly find the shortest path decreases.

\begin{figure}
    \centering
    \includegraphics[width=1\linewidth]{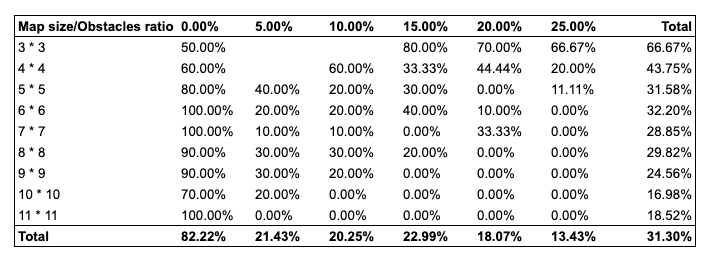}
    \caption{Result}
    \label{fig:enter-label}
\end{figure}

Additionally, we visualize the GPT-4's journal to find the path in 2D. The blue dotted lines represent the feasible path leading to the destination, and the red line denotes the path generated by GPT-4.

\begin{figure}[h!]
    \centering
    \includegraphics[width=1\linewidth]{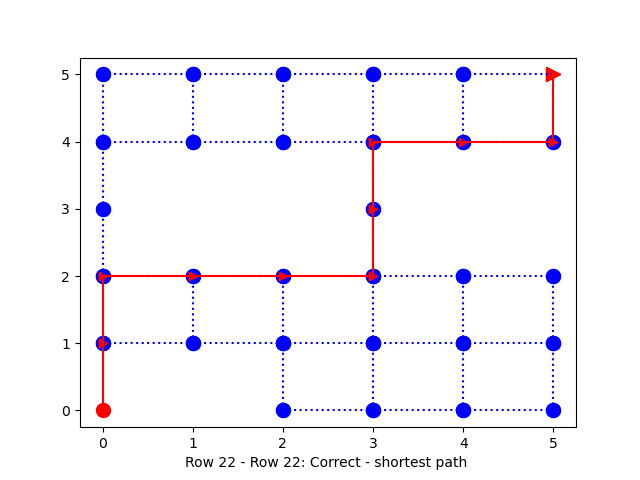}
    \caption{Example 1: 6$\times$6 grid with 15\% obstacles, GPT-4 correctly finds the shortest path}
    \label{fig:enter-label}
\end{figure}

\begin{figure}[h!]
    \centering
    \includegraphics[width=1\linewidth]{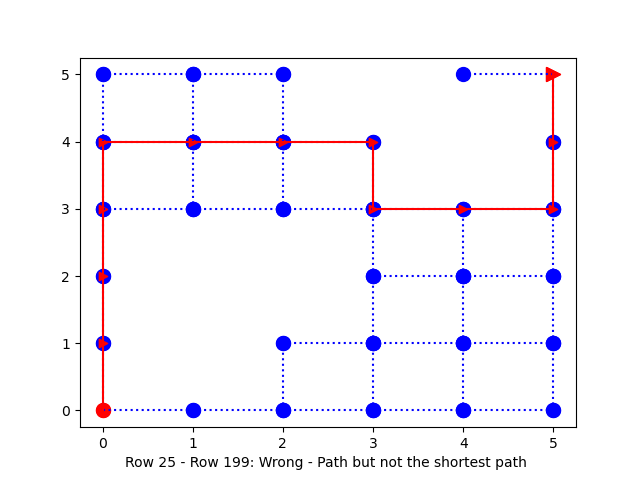}
    \caption{Example 2: 6$\times$6 grid with 15\% obstacles, GPT-4 does not find the shortest path because the total distance is not consistent with the Dijkstra algorithm }
    \label{fig:enter-label}
\end{figure}

\subsection{Experiments 2 - 3D Route Planning}
In our 3D pathfinding experiments, we evaluate the spatial reasoning abilities of GPT-4 by employing a similar strategy as in the 2D experiments to find the optimal path within a 3D grid. This involves extending the 2D space into 3D space, starting from the origin point $(0, 0, 0)$ and reaching the destination point $(x_{max}, y_{max}, z_{max})$. We still employ Dijkstra's algorithm as the benchmark to evaluate GPT-4's ability to find the shortest path in 3D.

\subsubsection{Results}
The chart shows GPT-4's accuracy in finding the shortest path in different map sizes and obstacle ratios in 3D. The results are similar to 2D pathfinding. As the map size and the obstacles ratio increase, the precision in finding the shortest path decreases.
By evaluating GPT-4's pathfinding abilities in both 2D and 3D space, we can conclude that GPT-4's ability to find the shortest path in a 2D grid is better than in a 3D grid.

\begin{figure}[h!]
    \centering
    \includegraphics[width=1\linewidth]{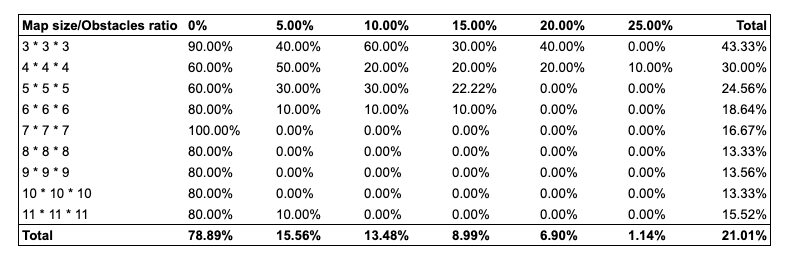}
    \caption{Result}
    \label{fig:enter-label}
\end{figure}

We also visualize the GPT-4's journal to find the path in 3D. The button left corner is the starting point$(0,0,0)$ and the top right corner is the destination point$(x_{max}, y_{max}, z_{max})$. The blue lines represent the available path leading to the destination, and the red line denotes the path generated by GPT-4.

\begin{figure}[h]
    \centering
    \includegraphics[width=1\linewidth]{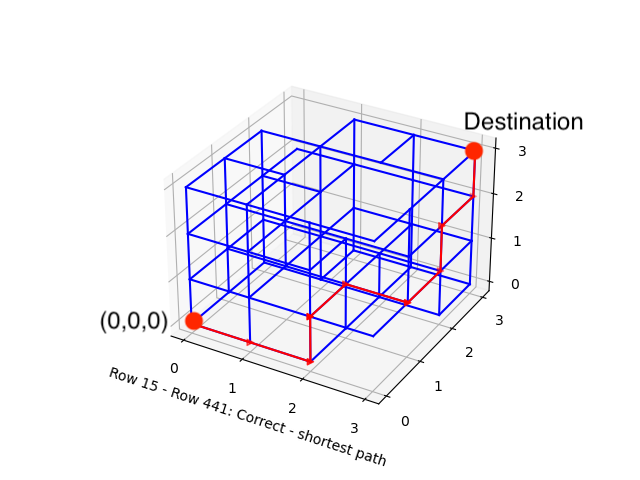}
    \caption{Example 1: 4$\times$4$\times$4 grid with 15\% obstacles, GPT-4 correctly finds the shortest path}
    \label{fig:enter-label}
\end{figure}
\begin{figure}[h]
    \centering
    \includegraphics[width=1\linewidth]{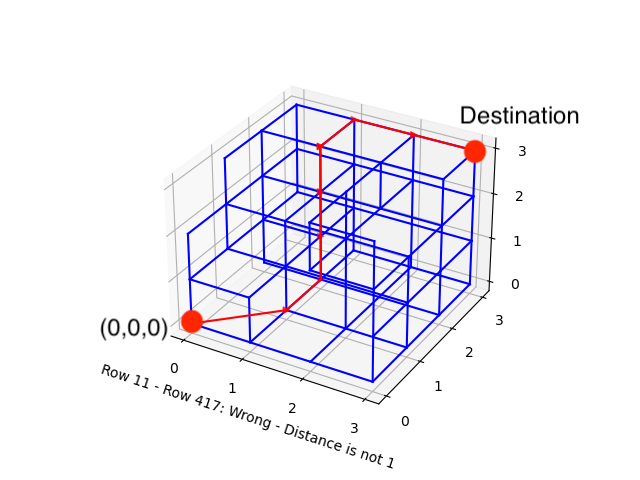}
    \caption{Example 2: 4$\times$4$\times$4 grid with 15\% obstacles, GPT-4 does not find the shortest path because GPT-4 does not adhere to the rule that restricts each step to 1 unit, so it is not a valid path}
    \label{fig:enter-label}
\end{figure}

\subsection{Experiments 3 - Plotting Spatial Points in 2D}

In this experiment, our objective is to evaluate GPT-4's ability to locate and plot points on a 2D map according to our text-based prompts.

\subsubsection{Experiment Setup}

To begin, we must establish benchmarking grids. These grids consist of dots,  with rows labeled with lowercase letters, and columns labeled by numbers. Then, we randomly mark some points on these grids by using uppercase letters. These benchmarking grids have 9 grid sizes, ranging from 2$\times$2 to 10$\times$10. Random uppercase letters are placed on the grids, with the letter density varying at 5\%, 10\%, 15\%, 20\%, and 25\%. Additionally, we generate 10 sets of random letters for each density. It's important to note that we only consider 25\%(1 random point) point ratio for 2x2 grids and 10\% and 20\% point ratios for 3x3 grids to ensure consistency in the experiment. GPT-4 will be tasked with plotting these points based on text-based prompts. In the end, we will compare GPT-4's results with the benchmark for evaluation.

\subsubsection{Results}

The chart shows GPT-4's accuracy in plotting the points in different map sizes and point ratios in 2D. The results reveal that GPT-4 has limited abilities to plot the points in large map sizes and large point ratios.

\begin{figure}[h!]
    \centering
    \includegraphics[width=1\linewidth]{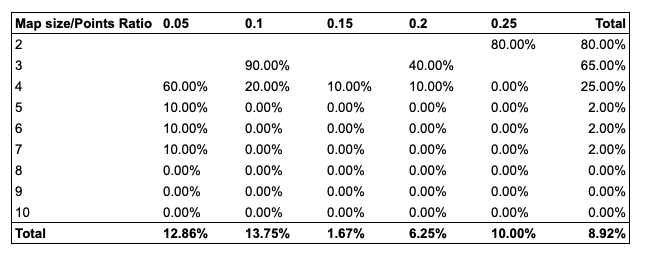}
    \caption{Enter Caption}
    \label{fig:enter-label}
\end{figure}

\begin{figure}[h!]
    \centering
    \includegraphics[width=1\linewidth]{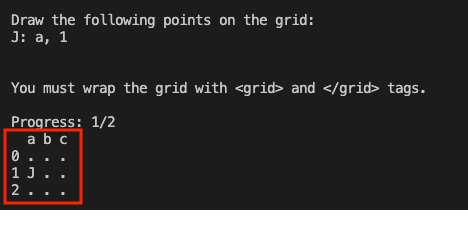}
    \caption{Example result by GPT-4: GPT-4 successfully find the correct point J}
    \label{fig:enter-label}
\end{figure}

\section{Conclusion}

In this paper, we introduced a novel framework and a custom dataset to evaluate the spatial reasoning abilities of large language models like GPT-4. Our findings provide important insights into the strengths and weaknesses of these models in understanding spatial information. By sharing our dataset and code, we aim to support further research in this area. This work not only advances our understanding of language models' capabilities in spatial reasoning but also lays the groundwork for future developments in technologies that require such skills.

\textbf{Limitation.} While our bespoke dataset is meticulously designed for this study, it might be too specific or not comprehensive enough to capture all aspects of spatial reasoning. This could limit the applicability of our findings to other spatial reasoning tasks or real-world scenarios.

\textbf{Future.} Future studies would develop more diverse and comprehensive datasets that include a wider range of spatial reasoning tasks, including dynamic and real-time spatial problems, to better understand the capabilities of language models in different scenarios.

\bibliography{custom}

\end{document}